\documentclass[10pt,twocolumn,letterpaper]{article}
\usepackage[pagenumbers]{arxiv}

\definecolor{myblue}{rgb}{0.21,0.49,0.74}
\usepackage[pagebackref,breaklinks,colorlinks,allcolors=myblue]{hyperref}
\usepackage{colortbl} 
\usepackage{array}    
\usepackage{graphicx} 
\usepackage{soul}
\usepackage{bm}
\usepackage[printonlyused]{acronym}

\title{PhysMotion: Physics-Grounded Dynamics from a Single Image}

\author{Xiyang Tan$^{1*}$ \quad Ying Jiang$^{1*}$ \quad Xuan Li$^{1*}$ \quad Zeshun Zong$^{1}$ \quad Tianyi Xie$^{1}$ \\ Yin Yang$^{2}$ \quad Chenfanfu Jiang$^{1}$ \\
$^{1}$ UCLA, $^{2}$ University of Utah}

\begin{document}

\newcommand\blfootnote[1]{%
  \begingroup
  \renewcommand\thefootnote{}\footnote{#1}%
  \addtocounter{footnote}{-1}%
  \endgroup
}

\acrodef{sds}[SDS]{Score Distillation Sampling}
\acrodef{vsd}[VSD]{Variational Score Distillation}
\acrodef{udf}[UDF]{Unsigned Distance Field}
\acrodef{cv}[CV]{Computer Vision}
\acrodef{cg}[CG]{Computer Graphics}
\acrodef{aigc}[AIGC]{Artificial Intelligence Generated Content}
\acrodef{vr}[VR]{Virtual Reality}
\acrodef{mr}[MR]{Mixed Reality}
\acrodef{nerf}[NeRF]{Neural Radiance Fields}
\acrodef{xpbd}[XPBD]{Extended Position Based Dynamics}
\acrodef{3dgs}[3DGS]{3D Gaussian Splatting}
\acrodef{vae}[VAE]{Variational Encoder}

\twocolumn[{%
\renewcommand\twocolumn[1][]{#1}%
\maketitle
\begin{center}
    \centering
    \captionsetup{type=figure}
    \includegraphics[width=\textwidth, trim=0 0 0 0, clip]{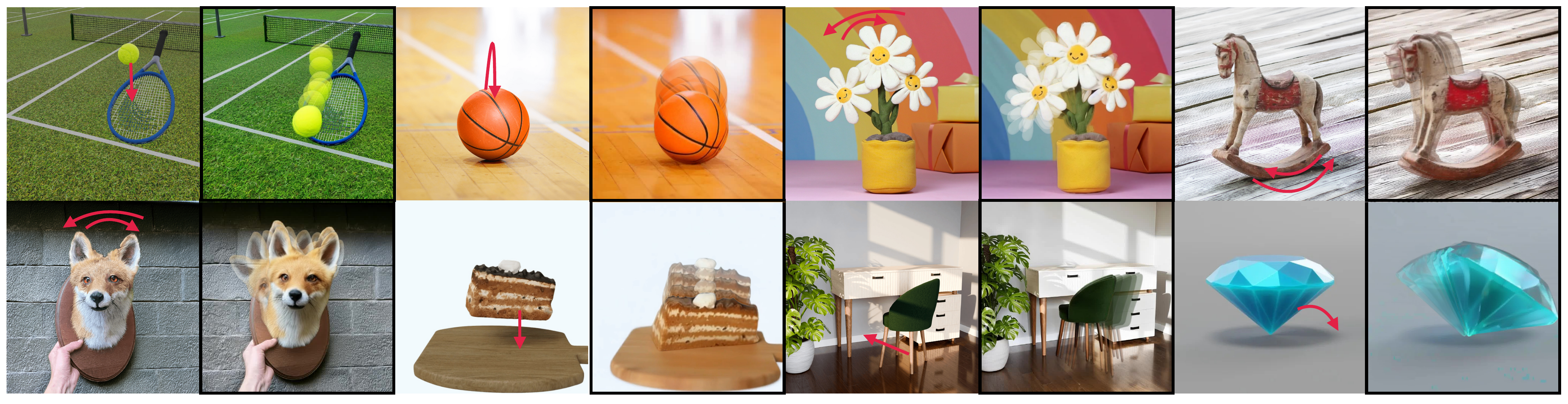}
    \captionof{figure}{
     \textbf{PhysMotion} is a novel image-to-video framework that uses physical dynamics and 3D geometric representations to generate realistic videos, capturing rigid body motion, elastic deformation, viscoplastic flow, and fracture. Each image pair shows the input image on the left and the generated dynamics on the right.
    }
\label{fig:teaser}
\end{center}
}]

\begin{abstract}

\blfootnote{* indicates equal contributions.}

We introduce PhysMotion, a novel framework that leverages principled physics-based simulations to guide intermediate 3D representations generated from a single image and input conditions (e.g., applied force and torque), producing high-quality, physically plausible video generation. By utilizing continuum mechanics-based simulations as a prior knowledge, our approach addresses the limitations of traditional data-driven generative models and result in more consistent physically plausible motions. Our framework begins by reconstructing a feed-forward 3D Gaussian from a single image through geometry optimization. This representation is then time-stepped using a differentiable Material Point Method (MPM) with continuum mechanics-based elastoplasticity models, which provides a strong foundation for realistic dynamics, albeit at a coarse level of detail. To enhance the geometry, appearance and ensure spatiotemporal consistency, we refine the initial simulation using a text-to-image (T2I) diffusion model with cross-frame attention, resulting in a physically plausible video that retains intricate details comparable to the input image. We conduct comprehensive qualitative and quantitative evaluations to validate the efficacy of our method. Our project page is available at: \url{https://supertan0204.github.io/physmotion_website/}.
\end{abstract}

\section{Introduction}
\label{sec:intro}
Video generation and editing are essential in movie and game industries, where achieving high-quality content efficiently still remains challenging. Traditional workflows with commercial software—such as video recording and post-production editing—are labor-intensive and require substantial expertise. These methods often result in huge time and labor costs and restrict accessibility for novices. Recent advances in deep learning, particularly in generative models, have enabled more automated and efficient approaches to video generation. State-of-the-art image and video diffusion models can synthesize videos from text prompts or image inputs. However, the inherent sparsity of text prompts often fails to capture detailed object layouts and complex motion features, resulting in generated videos that do not fully align with user intentions \cite{wang2024motionctrl, guo2025sparsectrl}. To support controllable video generation, additional conditions—such as motion trajectories, reference videos, poses, and camera movements—have been incorporated into diffusion models \cite{jeong2024vmc, shi2024motion, chen2023control, guo2025sparsectrl, zhang2024mimicmotion, he2023animate, yin2023dragnuwa}. Despite these advancements, diffusion models still struggle to maintain physical realism, often producing outputs that violate basic physical principles \cite{bansal2024videophy, meng2024towards}.

Previous work has employed physics simulators to generate physics-based dynamics and render videos through differentiable rendering techniques \cite{xie2023physgaussian, jiang2024vr, liu2025physgen, huang2024dreamphysics, zhang2025physdreamer, li2023pac, cai2024gaussian}. For example, PhysGaussian \cite{xie2023physgaussian}, VR-GS \cite{jiang2024vr}, and PhysGen \cite{liu2025physgen} utilize simulators such as MPM \cite{jiang2016material}, XPBD \cite{macklin2016xpbd}, and rigid body simulators \cite{blomqvist2023pymunk} to achieve realistic physics-driven video synthesis. To further streamline physics parameter tuning, PhysDreamer \cite{zhang2025physdreamer} distills video diffusion priors via \ac{sds} to estimate the physical material properties of static 3D objects, enabling the creation of immersive, realistic interactive dynamics. PAC-NeRF \cite{li2023pac} and GIC \cite{cai2024gaussian} estimate unknown geometry and physical parameters to ensure physically plausible outputs in \ac{nerf} and \ac{3dgs}, respectively. However, these techniques often require multi-view images, videos, or 3D objects as input, limiting their generalizability. While PhysGen \cite{liu2025physgen} generates dynamics from a single image, it is constrained to 2D dynamics. Our objective is to generate visually compelling, physics-grounded 3D dynamics from a single image, enhancing both visual quality and the physical plausibility of the resulting videos.

To address these challenges, we combine insights from both 3D and 2D generative models. Our goal is to generate realistic 3D dynamics, for which physical simulations are best performed on 3D geometry; therefore, we lift the image to a \ac{3dgs} representation. However, a direct application of \ac{3dgs} deformation and rendering yields low-quality videos. Inspired by Image Sculpting \cite{Yenphraphai2024imagesculpting}, we design a coarse-to-fine refinement stage that employs 2D image diffusion models and cross-attention mechanisms. This approach enables the final video to exhibit physically plausible motion consistent with 3D dynamics while preserving the quality and detail of the original input image. Our contributions include
\begin{itemize}
\item The first single image-to-video framework with 3D geometry awareness and physics-grounded dynamics.
\item A video enhancement module that improves video quality, maintains spatiotemporal consistency, preserves intricate details, and aligns closely with the input image.
\item Comprehensive evaluations on visual coherence, physical plausibility, and generation versatility.
\end{itemize}

\section{Related Work}
\label{sec:related}

\subsection{Generative Dynamics}
Previous works primarily follow two paradigms to generate dynamics from images or 3D content. Data-driven methods leverage image or video generative models to create dynamic scenes from various control signals, such as text prompts \cite{xing2025dynamicrafter, li2024animate}, bounding boxes \cite{li2024animate}, trajectories \cite{wang2024motionctrl, wu2025draganything, shi2024motion, zhang2024tora, li2024generative}, reference videos \cite{wei2024dreamvideo, zhao2023motiondirector, jeong2024vmc}, and camera movements \cite{wang2024motionctrl, yang2024direct}. They rely on pre-trained generative models, which often lack fundamental understanding of physics \cite{meng2024towards}. Consequently, the generated motions may violate physical laws.
In contrast, simulation-based methods produce generative dynamics that are physically grounded\cite{bansal2024videophy}. For instance, PhysGaussian \cite{xie2023physgaussian} and PhysDreamer \cite{zhang2025physdreamer} integrate a MPM simulator with \ac{3dgs} to generate physics-based dynamics. VR-GS \cite{jiang2024vr} employs \ac{xpbd} \cite{macklin2016xpbd} for real-time dynamics. \citet{zhong2025reconstruction} uses a spring-mass model within \ac{3dgs} to simulate elastic objects. These methods, however, require 3D contents or multi-view images as input. Recently, PhysGen \cite{liu2025physgen} utilizes rigid-body physics to create dynamics, but is limited to 2D rigid-body motions. In contrast, our work focuses on generating 3D deformable dynamics from a single image. Inspired by \cite{xie2023physgaussian}, we integrate MPM to enable physics-based dynamics, using only a single image as input. Unlike Phy124 \cite{lin2024phy124}, which outputs coarse video directly from the MPM simulator, ours produces higher-quality, visually compelling dynamics through video enhancement.

\subsection{Sparse-view 3D Reconstruction}
Recent advancements in neural rendering techniques, such as NeRF \cite{mildenhall2021nerf} and \ac{3dgs} \cite{kerbl3Dgaussians}, have made significant progress in 3D reconstruction. Approaches like additional depth supervision \cite{deng2022depth}, depth prior loss \cite{niemeyer2022regnerf, wang2023sparsenerf}, CLIP-based supervision \cite{jain2021putting}, pixel-aligned features \cite{yu2021pixelnerf}, frequency regularizer \cite{yang2023freenerf}, and diffusion priors \cite{wu2024reconfusion} have been introduced to promote sparse-view reconstruction. More relatedly, to reconstruct \ac{3dgs} scenes from sparse views,  methods \cite{szymanowicz24splatter, chen2024mvsplat, wewer2024latentsplat, paliwal2024coherentgs, xiong2023sparsegs, zhu2025fsgs, li2024dngaussian, charatan2024pixelsplat} have been developed to better constrain the optimization process. Among these, FSGS \cite{zhu2025fsgs} and SparseGS \cite{xiong2023sparsegs} apply monocular depth estimators or diffusion models to GS under sparse-view conditions. DNGaussian \cite{li2024dngaussian} uses both soft and hard depth regularization to restore accurate scene geometry.

\subsection{Diffusion Models}
\citet{sohl2015deep} first introduced diffusion models to learn a reverse diffusion process for data restoration. \citet{dhariwal2021diffusion} applied diffusion models to image synthesis, achieving higher visual quality in generated images than GANs. \citet{rombach2022high} proposed latent diffusion models, which first compress image data and then sample in a latent space during the denoising process. Besides image synthesis \cite{shi2024dragdiffusion, ge2023expressive, balaji2022ediff, podell2023sdxl, meng2023distillation, zhang2023adding, rombach2022high, dhariwal2021diffusion, kawar2023imagic, Yenphraphai2024imagesculpting}, diffusion models have been applied to many other tasks, including 3D reconstruction and generation \cite{gao2024cat3d, tang2023dreamgaussian, yu2024wonderworld, liu2023zero, long2024wonder3d, lin2023magic3d, poole2022dreamfusion, li2023instant3d}, video generation \cite{yang2024cogvideox, zhang2023i2vgen, wu2024aniclipart, tokenflow2023, qi2023fatezero, gu2024videoswap, ku2024anyv2v, liu2024video, yang2023rerender, chai2023stablevideo, liew2023magicedit, chen2024videocrafter2, menapace2024snap, liu2025physgen, wang2024motionctrl, shi2024motion}, robot policy \cite{chi2023diffusion, ke20243d, ze20243d}, and scene understanding \cite{lugmayr2022repaint, amit2021segdiff, chen2024mvip}. Our task focuses on enhancing video quality and ensuring alignment with input.
Previous works utilized pre-trained diffusion models as priors, or trained video diffusion models directly, to create and edit videos \cite{yatim2024space, ku2024anyv2v, harsha2024genvideo, gu2024videoswap, cohen2024slicedit, qi2023fatezero}. For example, \cite{yatim2024space} employed video diffusion models to learn space-time features for motion transfer; \cite{ku2024anyv2v} injected temporal features into frozen video diffusion models to edit video based on the first edited frame. However, due to the substantial computational and memory requirements, video diffusion models remain in its early stage, largely limited to short clips, or yielding low visual quality \cite{tokenflow2023}. Consequently, researchers leveraged image diffusion models, adapting them for video editing tasks through spatio-temporal slices \cite{cohen2024slicedit}, reference-aware latent correction \cite{harsha2024genvideo}, interactive semantic point maps \cite{gu2024videoswap}, or spatial-temporal attention \cite{qi2023fatezero}. Nonetheless, solely applying image diffusion models on video editing often causes flickering issues among frames \cite{chang2024warped}. To mitigate this, methods such as feature warping \cite{yang2023rerender, ni2023conditional}, cross-frame attention \cite{tokenflow2023, ceylan2023pix2video}, and noise warping \cite{chang2024warped} have been proposed to improve video consistency.

\section{Method}
\label{sec:method}

\begin{figure*}[t]
  \centering
  \includegraphics[width=1.0\textwidth]{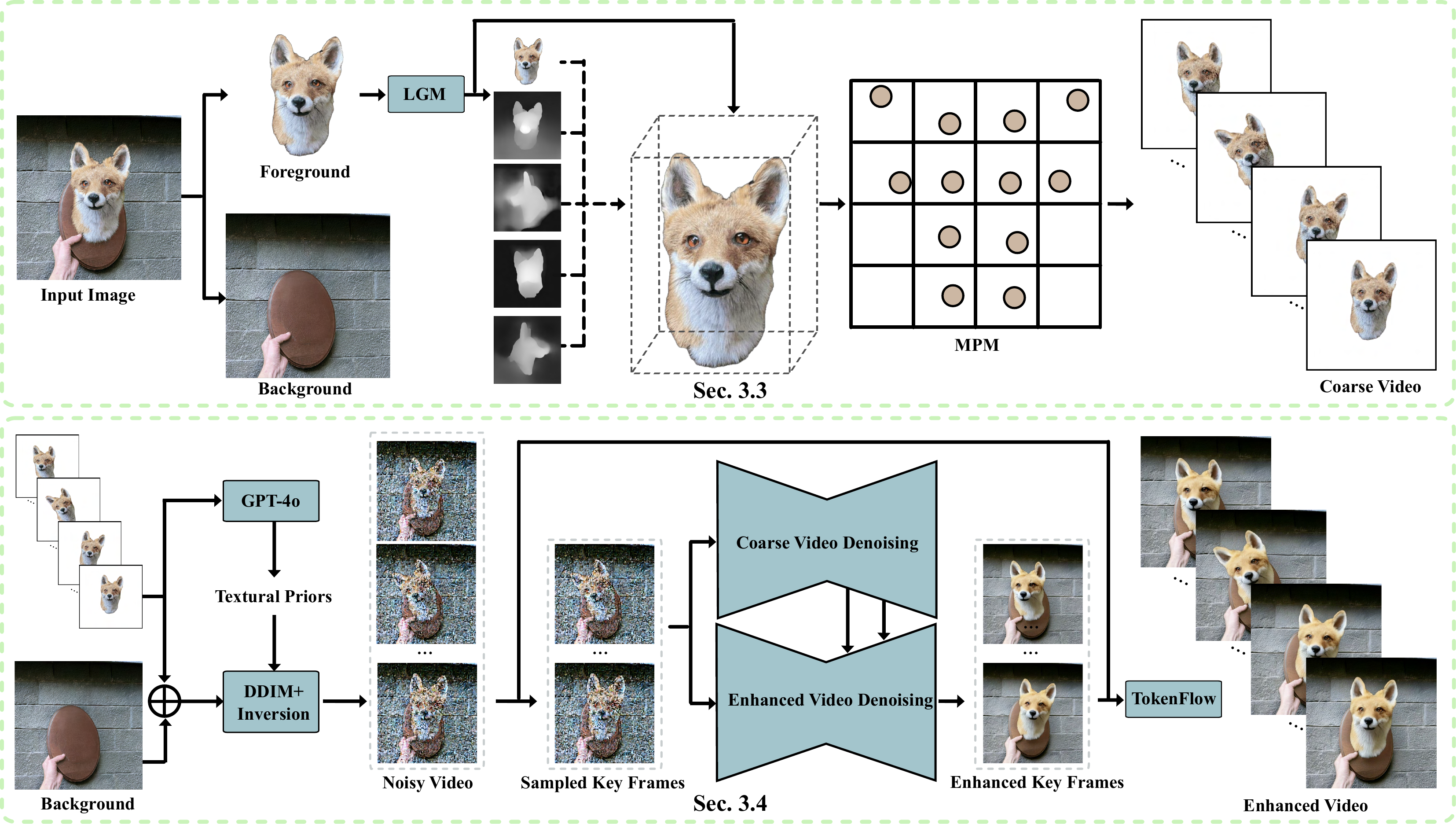}]
  \caption{
    \textbf{Overview.} Given a single image input $\mathbf{I}^0$, we introduce a novel pipeline to generate high-fidelity, physics-grounded video with 3D understanding. Our pipeline consists of two main stages: first, we perform a single-view 3DGS reconstruction of the segmented object from the input image, then synthesize a physics-grounded coarse object dynamics $\{\mathcal{I}_j\}_{j=1}^N$ (\cref{sec:geometry-aware reconstruction}); then, we apply diffusion-based video enhancement to produce the final enhanced video $\{\mathcal{I}^*_j\}_{j=1}^N$ with backgrounds (\cref{sec:generative-video-enhancement}), enabling users to create visually compelling, physics-driven video from a single image with an applied conditional force or torque.} 
  \label{fig:pipeline}
\end{figure*}

\subsection{Preliminaries}

\paragraph{3D Gaussian Splatting}
In its general format, 3D Gaussian Splatting (3DGS) \cite{kerbl3Dgaussians} represents a static scene \cite{mildenhall2021nerf} with a set of anisotropic 3D Gaussian kernels $G = \{\boldsymbol{\mathcal{G}}_k:(\mathbf{x}_k, \sigma_k, \mathbf{H}_k, \mathcal{C}_k)\}_{k\in \mathcal{K}}$
, where $\mathbf{x}_k$, $\sigma_k$, $\mathbf{H}_k$, and $\mathcal{C}_k$ are the centers, opacities, covariance matrices, and color representations of the Gaussians, respectively. At rendering stage, those 3D Gaussians are projected onto the 2D image space as 2D Gaussian kernels $\{\boldsymbol{\mathcal{G}}^{\text{proj}}_k\}_{k\in\mathcal{K}}$, and the color of each image pixel $\mathbf{p}$ is computed using
\begin{equation}
    \mathbf{C} (\mathbf{p}) = \sum_{n\in \mathcal{N}(\mathbf{p})} \alpha_n \mathbf{c}_n \prod_{j=1}^{n-1} (1-\alpha_j).
\end{equation}
Here $\mathcal{N}(\mathbf{p})$ denotes the Gaussians overlapping pixel $\mathbf{p}$; $\alpha_n$ represents the $z$-depth ordered effective opacities, \ie, products of the projected Gaussian weights $\boldsymbol{\mathcal{G}}_n^{\text{proj}}$ and their overall opacities $\sigma_n$; $\mathbf{c}_n$ is the decoded color for color representation $\mathcal{C}_n$. Given the camera center $\mathbf{o},$ we can further obtain the pixel-wise depth as
\begin{equation}\label{eq:3dgs depthmap}
    \mathbf{D}(\mathbf{p})=\sum_{n\in\mathcal{N}(\mathbf{p})}\|\mathbf{x}_n-\mathbf{o}\|_2\cdot\left({\alpha}_n\prod_{j=1}^{n-1}(1 - {\alpha}_j)\right).
\end{equation}

The end-to-end differentiable rendering enables efficient optimization of 3DGS parameters. As in \cite{xie2023physgaussian, zhang2025physdreamer, ren2023dreamgaussian4d, Wu_2024_CVPR}, dynamics are supported by making $\mathbf{x}_k, \mathbf{H}_k$ time-dependent.
 The rendering process is end-to-end differentiable which enables the optimization of 3DGS parameters.

\paragraph{MPM Simulation for Continuum Mechanics} 
In continuum mechanics, motion is prescribed by a continuous deformation map $\mathbf{x}=\boldsymbol{\varphi}(\mathbf{X},t)$ between the undeformed material space $\Omega^0\ni\mathbf{X}$ and the deformed world space $\Omega^t\ni\mathbf{x}$ at time $t$. The evolution of deformation map $\boldsymbol{\varphi}$ follows the conservation of mass and momentum \cite{li2024physics}. 
As essentially a Lagrangian method, the Material Point Method (MPM) naturally obeys conservation of mass. The conservation of momentum can be formulated as  $\rho(\mathbf{x},t) \dot{\mathbf{v}}(\mathbf{x},t) = \nabla\cdot\boldsymbol{\sigma}(\mathbf{x},t) + \mathbf{f}^{\text{ext}},$ where $\mathbf{v}(\mathbf{x},t)$ is the velocity field, $\rho(\mathbf{x},t)$ is the density field, $\mathbf{f}^{\text{ext}}$ is the external force per unit volume. $\bm \sigma = \frac1{\det(\mathbf{F})}\frac{\partial \Psi}{\partial \mathbf{F}}\mathbf{F} {\mathbf{F}}^\mathsf{T}$ is the Cauchy stress tensor associated with a energy density $\Psi(\mathbf{F})$, and $\mathbf{F}(\mathbf{X},t):=\nabla_{\mathbf{X}}\boldsymbol{\varphi}(\mathbf{X},t)$ is the local deformation gradient.

Various works have demonstrated MPM's superb ability and compatibility to endow 3DGS with physics-based dynamics \cite{xie2023physgaussian, huang2024dreamphysics, zhang2025physdreamer, lin2024phy124, ren2023dreamgaussian4d, Wu_2024_CVPR}. In MPM, the continuum is discretized by a set of particles, each representing a small material region. At each time step $t_n,$ we track their Lagrangian quantities such as position $\mathbf{x}_p$, velocity $\mathbf{v}_p$, and deformation gradient $\mathbf{F}_p$. 
To advance one time step, we discretize the momentum equation by forward Euler as
\begin{equation}
    \frac{m_i}{\Delta t} (\mathbf{v}^{n+1}_i - \mathbf{v}^{n}_i) = -\sum_p V_p^0 \frac{\partial \Psi}{\partial \mathbf{F}}\mathbf{F}^{ n}_p{\mathbf{F}^{ n}_p}^{\mathsf{T}} \nabla w_{ip}^n + \mathbf{f}^{\text{ext}}_i.
\label{eq:mpm_discretization}
\end{equation}
Here $i$ and $p$ represent the Eulerian grid and the Lagrangian particles, respectively; $w_{ip}^n$ is the B-spline kernel defined on $i$-th grid evaluated at $\mathbf{x}_p^n$; $V_p^0$ is the initial representing volume, and $\Delta t$ is the time step size. The updated grid velocity field $\mathbf{v}_i^{n+1}$ is transferred back to particles as $\mathbf{v}_p^{n+1}$, updating the particles' positions to $\mathbf{x}_p^{n+1} = \mathbf{x}_p^n + \Delta t \mathbf{v}_p^{n+1}$.  We refer to \cite{zong2023neural, li2022energetically, zong2024convex} for more discussions on plasticity handling, and \cite{jiang2016material, stomakhin2013mpmsnow, gast2015optimization} for more details on MPM.

\paragraph{Physics-Integrated 3D Gaussians} We follow \cite{xie2023physgaussian} to generate physics-grounded dynamics of 3DGS. Applying a first-order approximation of the deformation map, continuum mechanics is combined with 3DGS by
$\mathbf{x}\approx\tilde{\boldsymbol{\varphi}}_k(\mathbf{X},t):=\mathbf{x}_k+\mathbf{F}_k(\mathbf{X}-\mathbf{X}_k)$, and the Gaussian distribution function for each Gaussian kernel $p$ is augmented as, $\forall\mathbf{x}\in\mathbb{R}^3,$ \begin{equation}
\boldsymbol{\mathcal{G}}_k(\mathbf{x})=\exp\left\{-\frac{1}{2}(\mathbf{x}-\mathbf{x}_k)^{\mathsf{T}}(\mathbf{F}_k\mathbf{H}_k\mathbf{F}_k^{\mathsf{T}})(\mathbf{x}-\mathbf{x}_k)\right\}.
\end{equation}This transformation gives a time-dependent 3DGS geometrical framework:\begin{equation}
    \mathbf{x}_k(t)=\boldsymbol{\varphi}(\mathbf{X}_k,t),\quad \mathbf{h}_k(t)=\mathbf{F}_k(t)\mathbf{H}_k\mathbf{F}_k(t)^{\mathsf{T}}.
\end{equation}
The world-space covariance matrix $\mathbf{h}_k$ can be subsequently updated as \begin{equation}
\mathbf{h}_k^{n+1}=\mathbf{h}_k^n+\Delta t\left(\nabla\mathbf{v}_k\mathbf{h}_k^n+\mathbf{h}_k^n\nabla\mathbf{v}_k^\mathsf{T}\right).
\end{equation}
In this work, we use only first order sphere harmonics (SH) \cite{kerbl3Dgaussians}. As color is not view-dependent with order one SH, the polar decomposition to a rotation matrix is not needed \cite{xie2023physgaussian}.

\subsection{Pipeline Overview}
Given a single image as input, our pipeline mainly consists of two parts: 1) we extract coarse 3DGS representation of foreground object, followed by applying a geometry-aware optimization. We generate physics-grounded dynamics of object using a MPM simulator; 2) we apply a diffusion-based video enhancement pipline to generate high-fidelity video with realistic object-background interaction.

\subsection{Geometry-Aware Reconstruction}
\label{sec:geometry-aware reconstruction}
Starting from a single image $\mathbf{I}^0$, using SAM \cite{kirillov2023segany}, we segment the image as a foreground $\mathbf{I}^0_F$ containing the object of interest, and a background $\mathbf{I}^0_B$. 

A spatially consistent 3DGS representation of the foreground object is pivotal to high-fidelity physics-based dynamics.  We use LGM \cite{tang2024lgm} to obtain a coarse initial 3DGS reconstruction for the foreground object. We then optimize the coarse 3DGS with depth and color supervision. Motivated by \cite{li2024dngaussian}, we include a hard-depth loss during the optimization stage. Given $\mathbf{I}^0_F,$ as done in \cite{tang2024lgm}, MVDream \cite{shi2023MVDream} is used to generate four multi-view images $\{\hat{\mathbf{I}}_{F}^0, \hat{\mathbf{I}}_F^1, \hat{\mathbf{I}}_{F}^2,\hat{\mathbf{I}}_{F}^3\}$ , where $\hat{\mathbf{I}}_{F}^a$ is the image generated at viewpoint with azimuth $\theta = \pi a/2$ for $a = 0, 1, 2,$ and $3$. Each generated image $\hat{\mathbf{I}}_{F}^a$ is decomposed as a set of pixel patches $\hat{\mathbf{I}}_{F}^a = \bigcup\limits_i \mathbf{P}_i^{a}$. The hard-depth loss $ \mathcal{L}_{\text{hard}}$ for each pixel patch is thus formulated as 
\begin{equation}
    \mathcal{L}_{\text{hard}}(\mathbf{P}_i^{a}) = \left\| \text{N}\left(\mathbf{D}_{\text{hard}}(\mathbf{P}_i^{a}) - \hat{\mathbf{{D}}}(\mathbf{P}_i^{a})\right)\right\|_2,
\end{equation}
where hard-depth $\mathbf{D}_{\text{hard}}(\mathbf{P}_i^{a})$ is defined as 
\begin{equation}
    \mathbf{D}_{\text{hard}}(\mathbf{P}_i^{a}) = \sum_{\mathbf{p} \in \mathbf{P}_i^{a}}  \sum_{n\in\mathcal{N}(\mathbf{p})}\|\mathbf{x}_n-\mathbf{o}_a\|_2\cdot\left(\delta(1-\delta)^{p-1}\boldsymbol{\mathcal{G}}_n^{\text{proj}}(\mathbf{p})\right).
\end{equation}
$\boldsymbol{\mathcal{G}}_n^{\text{proj}}(\mathbf{p})$ represents the projected 2D Gaussians overlapping pixel $\mathbf{p}$; $\mathbf{o}_a$ is the camera center of view $a;$ $\delta \in (0,1)$ is a scalar closer to $1$ so that the camera only sees the nearest Gaussians. $\hat{\mathbf{D}}(\cdot)$ is the ground truth monocular depth extracted from the training image pixel patch; $\text{N}(\cdot)$ is a balanced local and global normalization operator as in \cite{li2024dngaussian}.

The final hard-depth loss can be written as 
\begin{equation}
    \mathcal{L}_{\text{hard}} = \sum_{a=1}^3 \sum_{\mathbf{P}_i^{a} \subset \hat{\mathbf{I}}^a_F} \mathcal{L}_{\text{hard}}(\mathbf{P}_i^{a}) + \sum_{\mathbf{P}_i^{0} \subset \mathbf{I}^0_{F}} \mathcal{L}_{\text{hard}}(\mathbf{P}_i^{0}),
\end{equation}
where we replace $\hat{\mathbf{I}}^0_{F}$ with the available ground truth image $\mathbf{I}^0_{F}.$ Following \cite{li2024dngaussian}, we freeze all parameters except $\mathbf{x}_k, k\in\mathcal{K}$ for the depth supervision, thus ensuring the optimization of geometrical information.

For color supervision, we observe that using generated multi-view images results in poor spatial consistency, due to the inaccurate camera poses. Hence, color supervision is only applied on the input view image $\mathbf{I}^0_{F}.$ Following \cite{kerbl3Dgaussians}, we write the color supervision as \begin{equation}
\mathcal{L}_{\text{color}}=\mathcal{L}_1\left(\mathbf{I}(G,0),\mathbf{I}^0_{F}\right)+\lambda\mathcal{L}_{\text{D-SSIM}}\left(\mathbf{I}(G,0),\mathbf{I}^0_{F}\right),
\end{equation}
where $\mathbf{I}(G,0)$ is the rendered image at view with azimuth $\theta=0$ from the Gaussians $G=\{\boldsymbol{\mathcal{G}}_k\}_{k\in\mathcal{K}}$. 
The output of the optimized Gaussians are send to a MPM simulator, resulting in a coarse object dynamics $\{\mathcal{I}_j\}_{j=1}^N$ consisting of $N$ frames. We adopt the open-sourced MPM solver \cite{zong2023neural}. 
 \begin{figure}
    \includegraphics[width=\linewidth]{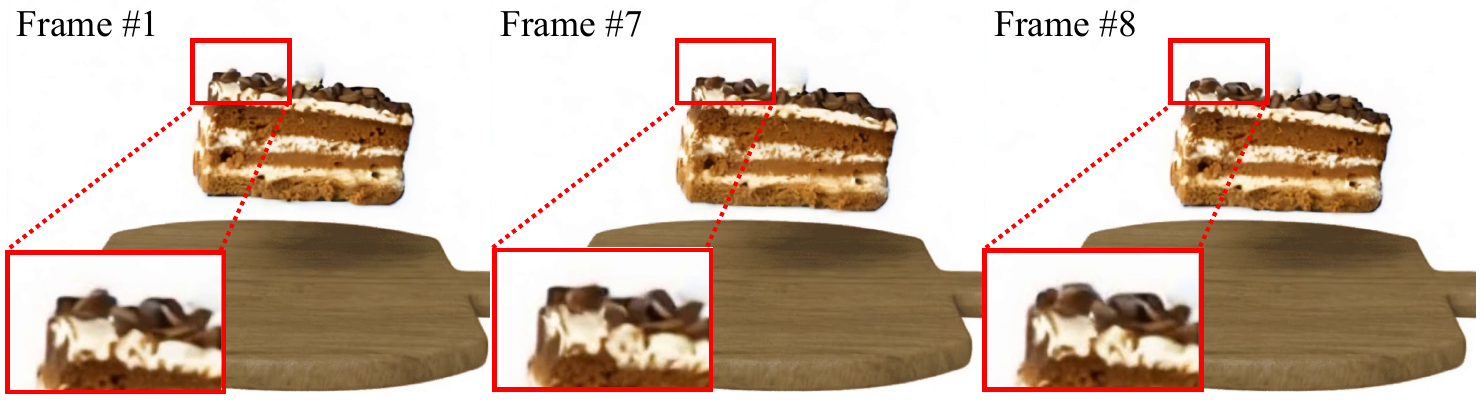}
    \caption{Applying blending in latent space causes temporal inconsistency at cake's edge. }\label{fig:mask_compare}
  \end{figure}
\subsection{Generative Video Enhancement}
\label{sec:generative-video-enhancement}
We present a diffusion-based video enhancement pipeline to generate enhanced video $\{\mathcal{I}^*_j\}_{j=1}^N$ from the coarse frames $\{\mathcal{I}_j\}_{j=1}^N$, which integrates object dynamics with the background $\mathbf{I}_{B}^0$  for realistic interactions, and encourages generative dynamics to maintain temporal consistency. Inspired by \cite{tokenflow2023}, given the coarse frames $\{\mathcal{I}_j\}_{j=1}^N$, a subset of key-frames $\{\mathcal{I}_{j_k}\}_{k=1}^K\subset \{\mathcal{I}_j\}_{j=1}^N$, with $K<N,$ is selected for key-frame enhancement. The key-frames $\{\mathcal{I}_{j_k}\}_{k=1}^K$ are first jointly enhanced with shared attention features, then the enhanced features of key-frames are propagated through the whole video to ensure temporal consistency. In our setting, the first frame is always selected as a key frame, \ie $j_1\equiv1$, as it contains the most faithful information to the input $\mathbf{I}^0$. We introduce our video enhancement pipeline in two stages: inversion stage (\cref{subsec:inversion stage}) and sampling stage (\cref{subsec:sampling stage}).

\begin{figure*}[htbp]
 \includegraphics[width=\linewidth]{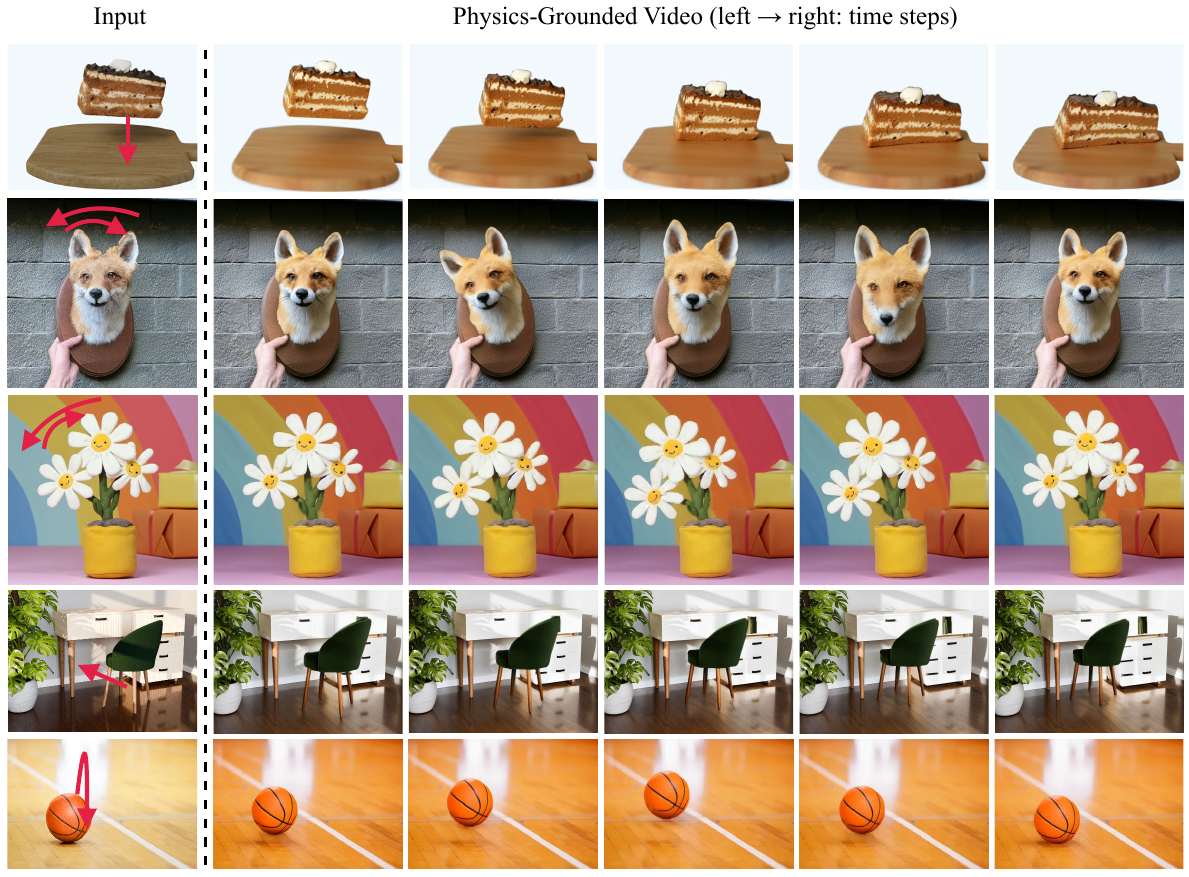}
  \centering
    \caption{\textbf{Showcases.} We demonstrate exceptional versatility of our approach across a wide variety of examples.}
  \label{fig:results}
\end{figure*}
\begin{figure*}[htbp]
 \includegraphics[width=1.0\textwidth]{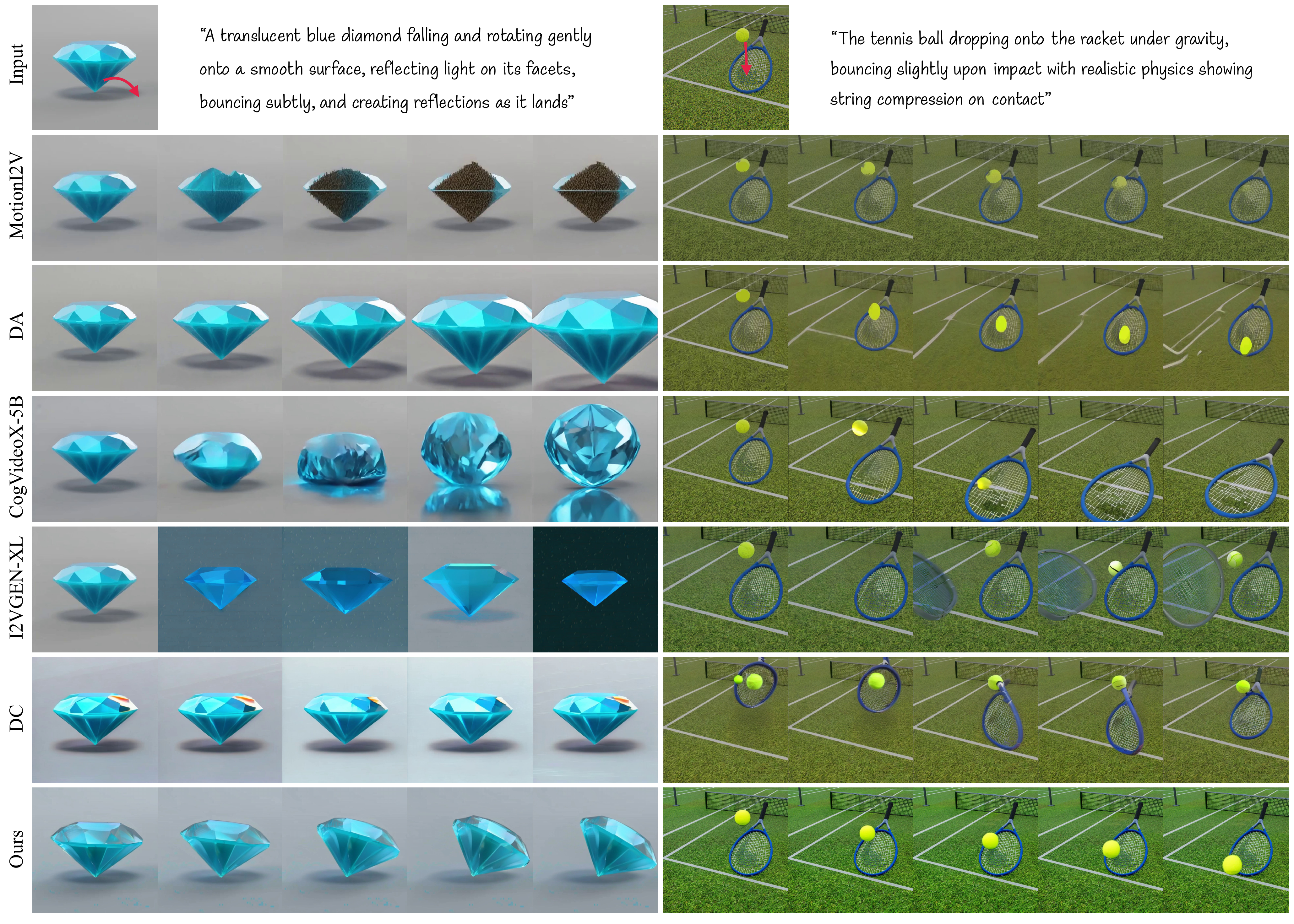}
  \centering
    \caption{\textbf{Qualitative Comparison}. We use compare our results against MotionI2V \cite{shi2024motion}, DragAnything (DA) \cite{wu2025draganything}, CogVideoX-5B \cite{yang2024cogvideox}, DynamiCrafter (DC) \cite{xing2025dynamicrafter} and I2VGen-XL \cite{shi2024motion}. Text prompts for CogVideoX-5B, I2VGen-XL and DynamiCrafter are generated using ChatGPT-4o, while trajectories are used for DragAnything and Motion-I2V.}
  \label{fig:comparisons}
\end{figure*}

\subsubsection{Inversion Stage}\label{subsec:inversion stage}
We use the $\bm\alpha$ values from the optimized foreground 3DGS to obtain blended image $\Bar{\mathcal{I}}_{j} = \bm\alpha \mathcal{I}_{j} + (\mathbf{1} - \bm\alpha) \mathbf{I}^0_{B}, 1\leq j \leq N$.
A DDIM+ inversion \cite{Yenphraphai2024imagesculpting} is applied on blended images to obtain the noisy latents $\{\tilde{\mathcal{I}}_{j}\}_{j=1}^N$ as \begin{equation}
   \{\bar{\mathcal{I}}_j\}_{j=1}^N \xrightarrow[\text{Inversion}]{\text{DDIM+}} \{\tilde{\mathcal{I}}_j\}_{j=1}^N.
\end{equation}
They capture the intricate appearance of the object and spatial information, enhancing the realistic generative dynamics with high-quality texture and geometric details.

The DDIM+ procedure is outlined as the following. We fine-tune a pre-trained image diffusion model \cite{ruiz2022dreambooth} based on the segmented object and generated multi-view images. Leveraging image diffusion priors, this process enhances the consistency of texture and appearance of the object with input image. Additionally, we utilize the depth map $\mathbf{D}(\mathcal{I})$ rendered from 3DGS and the canny edge $\mathbf{C}(\mathcal{I})$ as control signals for ControlNet \cite{zhang2023adding} to maintain geometric consistency between the foreground and the background within each frame. Moreover, textural priors are extracted by GPT-4o \cite{achiam2023gpt} to further augment the DDIM+ inversion process.

It could be noted that, unlike \cite{Yenphraphai2024imagesculpting}, we blend the foreground and background in image space before DDIM+ inversion, rather than in latent space at the sampling stage. Although blending in latent space preserves the original background and ensures output consistency with the input image, it heavily depends on the accuracy of the mask. Misalignments during mask encoding can lead to boundary inconsistencies in the final video, as illustrated in \cref{fig:mask_compare}
, making it unsuitable for video enhancement tasks. 
\subsubsection{Sampling Stage}\label{subsec:sampling stage}
During DDIM+ sampling stage, we perform coarse and enhanced sampling processes simultaneously. Following \cite{Tumanyan_2023_Pnp}, we switch the output of residual blocks and self-attention blocks in the enhanced sampling stage with corresponding outputs from the coarse sampling stage as $\forall j, 1\leq j\leq N:$
\begin{align*}
    \mathbf{f}_e(\tilde{\mathcal{I}}_j) &\leftarrow \mathbf{f}_c(\tilde{\mathcal{I}}_j), \\
    (\mathbf{Q}_{e}(\tilde{\mathcal{I}}_j), \mathbf{K}_{e}(\tilde{\mathcal{I}}_j)) &\leftarrow (\mathbf{Q}_{c}(\tilde{\mathcal{I}}_j), \mathbf{K}_{c}(\tilde{\mathcal{I}}_j)).
\end{align*}
The subscripts $e$ and $c$ stand for the features in \textbf{e}nhanced stage and \textbf{c}oarse stage;  $\mathbf{f}_{*}(\tilde{\mathcal{I}}_j),*=e,c$ indicates the output of residual blocks for the latent frame $\tilde{\mathcal{I}}_j$; $\mathbf{Q}_*$ and $\mathbf{K}_*$, $*=e,c$ are queries and keys within the transformer attention blocks. Feature injection is applied to all upsampling layers (\ie the decoding stage) in the UNet. The timesteps for feature and attention injection is controlled by two hyperparameters, $\tau_f$ and $\tau_A \in (0,1)$. 

For latent key-frames $\{{\tilde{\mathcal{I}}}_{j_k}\}_{k=1}^K$ during enhanced sampling, the self-attention features $\{\mathbf{Q}_{j_k}\}_{k=1}^K$ (queries), $\{\mathbf{K}_{j_k}\}_{k=1}^K$ (keys), $\{\mathbf{V}_{j_k}\}_{k=1}^K$ (values) are concatenated and shared to form the extended attention, with the queries and keys being replaced by the corresponding values in the coarse stage (for simplicity, we denote the $\mathbf{Q}_*(\tilde{\mathcal{I}}_{j})$ as $\mathbf{Q}_{j,*}$, where $*=e,c$, same as $\mathbf{K}$) as, for $k = 1, 2, ..., K,$
\begin{equation}\label{eqn:shared attention}
    \mathcal{A}_{j_k,e}=\text{Softmax}\left(\frac{\mathbf{Q}_{j_k,c}[\mathbf{K}_{j_1,c},\cdots,\mathbf{K}_{j_K,c}]^{\mathsf{T}}}{\sqrt{d}}\right).\;
\end{equation}
Here $d$ is the dimension of embedded vectors within $\mathbf{Q},\mathbf{K}$ \cite{Vaswani2017attn}. For each latent key-frame $\tilde{\mathcal{I}}_{j_k},$ its output of attention block during enhanced sampling $\bm\phi_e(\tilde{\mathcal{I}}_{j_k})$ is thus \begin{equation}\label{eqn:output of attention}
\bm\phi_e(\tilde{\mathcal{I}}_{j_k})=\mathcal{A}_{j_k,e}\cdot[\mathbf{V}_{j_1,e},\cdots,\mathbf{V}_{j_K,e}], \;\forall 1\leq k\leq K.
\end{equation}

For latent non-key-frames, following \cite{tokenflow2023}, we propagate the enhanced key-frames to them, through the extracted Nearest-Neighbor correspondences from the coarse blended video $\{\bar{\mathcal{I}}_j\}_{j=1}^N$. For a spatial location $\mathbf{q}$ in the feature map of frame $j\in\{1, \cdots, N\}\backslash\{j_1,\cdots,j_K\}$ , we replace its attention block output by a weighted average between neighboring key-frames
\begin{equation}
    \bm\phi_e(\tilde{\mathcal{I}}_j,\mathbf{q})=w_j\bm\phi_e(\tilde{\mathcal{I}}_{j+})\left[\nu_{j+}[\mathbf{q}]\right]+
    (1-w_j)\bm\phi_e(\tilde{\mathcal{I}}_{j-})\left[\nu_{j-}[\mathbf{q}]\right].
\end{equation}

Here, $j\pm$ denotes the index of the closest future (+) and past (-) key frames; $w_j\in(0,1)$ is a scalar proportional to the distance between frame $j$ and its neighboring key frames; $\nu_{j\pm}$ is defined over coarse stage on the original blended video:\begin{equation}
    \nu_{j\pm}[\mathbf{q}]=\mathop{\mathrm{arg\,min}}_{\mathbf{q}^*}\mathcal{D}\left(\bm\phi_c(\bar{\mathcal{I}}_j)[\mathbf{q}],\bm\phi_c(\bar{\mathcal{I}}_{j\pm})[\mathbf{q}^*]\right),
\end{equation}
where $\mathcal{D}$ is the cosine distance. The entire enhanced sampling will output a enhanced video:
\begin{equation}
    \{\tilde{\mathcal{I}}_{j}\}_{j=1}^N \rightarrow \{\bm\phi_{j,e}\}_{j=1}^N;\{\mathbf{f}_{j,e}\}_{j=1}^N\rightarrow \{\mathcal{I}_j^*\}_{j=1}^N.
\end{equation}

\section{Experiment}
\label{sec:exp}
\subsection{Results}

\paragraph{Implementation Details}  
We use MiDaS \cite{stan2023ldm3d} for monocular depth extraction for 3DGS optimization. For diffusion-based video enhancement, we leverage a pre-trained Stable-Diffusion \cite{Rombach_2022_CVPR} and ControlNet \cite{zhang2023adding} weights, and we fine-tune them using LoRA \cite{hu2022lora} trained for $1500$ steps with learning rate $10^{-6}$.  Detailed parameters and model settings are provided in the Appendix.

\paragraph{Showcases} We showcase various 3D-generated dynamics produced by PhysMotion (\cref{fig:results}). Our method creates visually compelling, temporal-consistent videos with physics-grounded dynamics. PhysMotion supports material properties such as elasticity, plasticity, rigid bodies, granular particles, and fracture effects. Additionally, the method effectively handles collisions and multi-object interactions.

\subsection{Quantitative Evaluation}
\paragraph{Baselines and Metrics}
For quantitative evaluation, we compare PhysMotion with several open-sourced state-of-the-art image-to-video generation models: I2VGen-XL \cite{zhang2023i2vgen}, CogVideoX-5B \cite{yang2024cogvideox}, DynamiCrafter \cite{xing2025dynamicrafter}, MotionI2V \cite{shi2024motion}, and DragAnything \cite{wu2025draganything}. The first three methods use text-based conditions, while the latter two use trajectory-based conditions to generate dynamics from a single image. Following \cite{shi2024motion},  we build a test set that includes 13 images of different scenes covering various physical materials. We use ChatGPT-4o \cite{achiam2023gpt} to generate the prompts for each image on possible dynamics for video generation. For a fair comparison, input images are resized to the desired height-width ratio of each model. The final output videos are resized to $768\times768$, same as the output of our pipeline.   
We adopt three different metrics: (1) the Physical Commonsense (PC) of generated videos reported by VideoPhy \cite{bansal2024videophy}; (2) the Semantic Adherence (SA) between videos and text prompt reported by VideoPhy; and (3) the temporal consistency (TC), physics plausibility (PP), and user preference (UP), assessed through a user study. 
As VideoPhy gives scores at different scales on different scenes, we apply a standard $z$-score normalization and report the $z$-score averaged over all scenes for a meaningful comparison. The normalized $z$-scores for each scene have mean zero and variance of one. See the Appendix for more details.

For user study, we adopt a two-alternative forced choice (2AFC)) method as suggested by \cite{tokenflow2023, park2020swapping, kolkin2019style}, where participants are asked to select which video demonstrates greater temporal consistency, visual quality, and adherence to physics in motion among two results generated from ours and a baseline respectively. The survey includes approximately 1,000 judgments per baseline, provided by 20 users. 
\begin{table}[t]
\centering
\caption{\textbf{Quantitative Comparisons.} We use VideoPhy \cite{bansal2024videophy} to compare the average normalized Physical Commonsense (PC) and Semantic Adherence (SA) scores, along with user study results showing preferences for our method over baselines in visual quality (VQ), temporal consistency (TC), and overall preference.} 
\setlength\tabcolsep{1.8pt}
\label{tab:quantitative comparison}
\small{
\begin{tabular}{p{0.9in}p{0.5in}p{0.5in}p{0.3in}p{0.3in}p{0.4in}}

\hline
\textbf{Methods} & \textbf{PC $\uparrow$} & \textbf{SA $\uparrow$} &  \textbf{TC} & \textbf{PP} & \textbf{Overall} \\ \hline  
I2VGen-XL  & 0.1853 & -0.8446  & 89\% & 90\% & 90\%\\ 
CogVideoX-5B & 0.0692 & 0.3480 & 76\% & 76\% & 78\% \\
DynamiCrafter & -0.2337 & -0.3091 & 88\% & 92\% & 92\%\\ 
MotionI2V & -0.5896 & 0.1530 & 92\% & 94\% & 91\%\\ 
DragAnything & 0.0545 & -0.0703 & 92\% & 90\% & 93\%\\ 
\hline
Ours & \cellcolor{Apricot}{0.5142} & \cellcolor{Apricot}{0.7230} & \textemdash & \textemdash & \textemdash \\ 
\hline
\end{tabular}
}
\end{table}
\paragraph{Comparison}
As indicated in \cref{tab:quantitative comparison},  based on the Physics Commonsense (PC) and Semantic Adherence (SA) evaluation, PhysMotion achieved the highest $z$-scores in both directions, indicating superior performance across various physics scenes. These results validate the effectiveness of PhysMotion in generating accurate dynamics with excellent semantic adherence across diverse scenarios. 
Furthermore, our proposed method consistently outperforms all baselines across every evaluation criterion in the user preference results (in both TC, PP, and overall). These results demonstrate PhysMotion's capability to synthesize realistic physical dynamics while maintaining temporal consistency.

\subsection{Qualitative Evaluation}
We show the qualitative comparison results in \cref{fig:comparisons}. MotionI2V \cite{shi2024motion} tends to shift textures without synthesizing true geometric motion, resulting in shape inconsistencies. DragAnything \cite{wu2025draganything}, often misinterpreting the motion relationship between the camera and object, and moves the camera instead of the object. Image-to-video generative methods with text conditions, such as DynamiCrafter \cite{xing2025dynamicrafter}, I2VGen-XL \cite{zhang2023i2vgen}, and CogVideoX-5B \cite{yang2024cogvideox}, frequently produce inconsistent motion or unintended object deformations. For example, in I2VGen-XL and CogVideoX-5B, the tennis ball disappears, and in DynamiCrafter, the tennis racket deforms and incorrectly embeds the tennis ball. In contrast, our physics-guided approach ensures physically plausible dynamics and high visual quality, preserving both the geometry and texture details of the input image.

\begin{figure}
    \includegraphics[width=\linewidth]{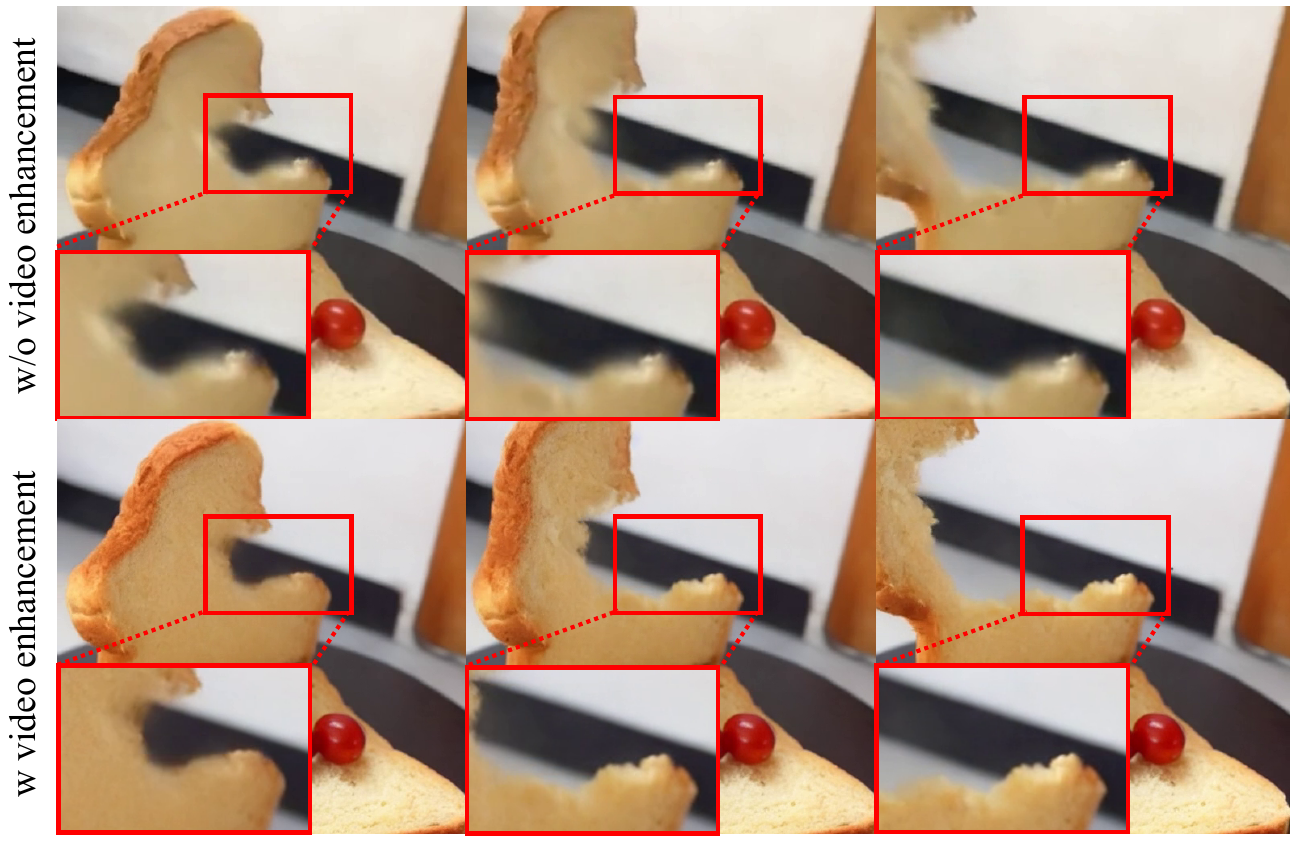}
    \caption{\textbf{Video Enhancement.} Our generative video enhancement captures the realistic texture of bread in the torn area.}\label{fig:ablation3}
\end{figure}
\subsection{Ablation Studies}
\begin{figure}
    \includegraphics[width=\linewidth]{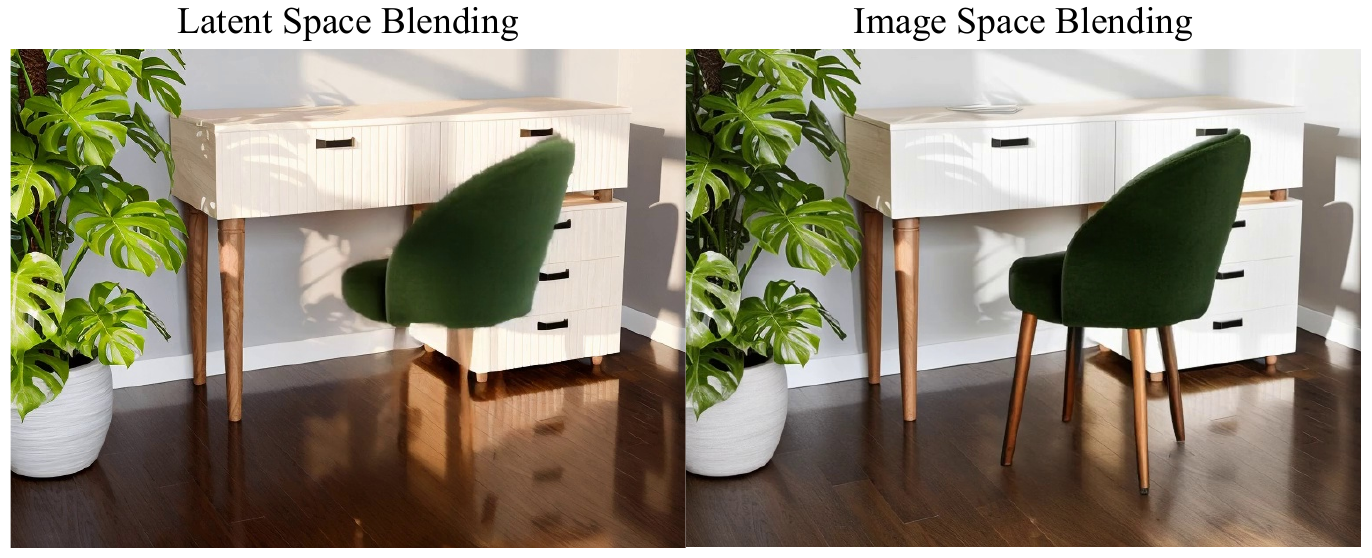}
    \caption{Background blending in latent space yields poor results due to inaccurate masking compared to blending in image space.}\label{fig:ablation2}
  \end{figure}

\begin{figure}
 \includegraphics[width=\linewidth]{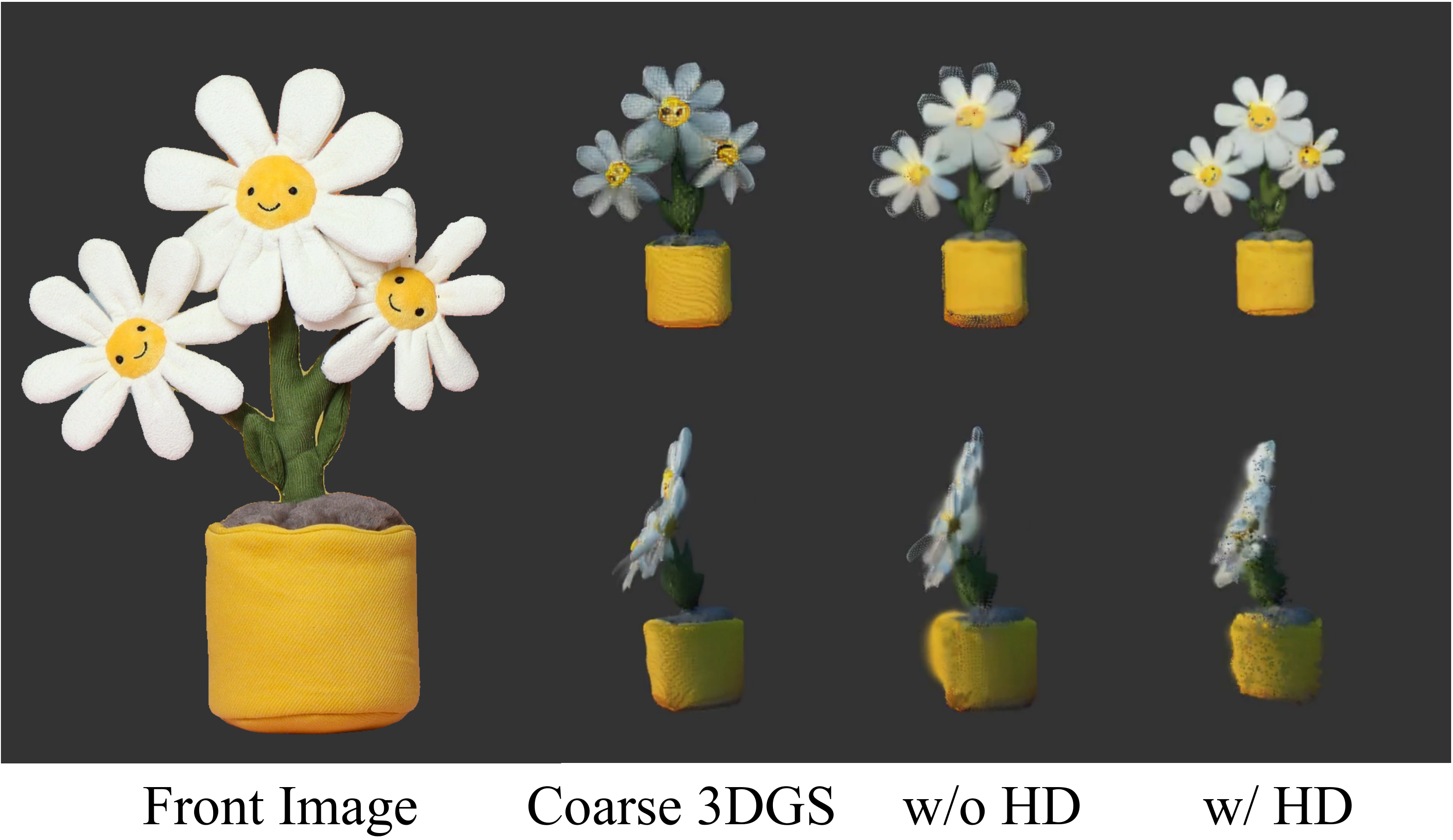}
    \caption{Coarse 3DGS shows blur and texture errors; without hard-depth (HD) supervision, floaters appeared; with hard-depth (HD) supervision, the reconstruction displays better appearance and correct shape.}
  \label{fig:ablation1}
\end{figure}

We conduct ablation studies to justify the necessity of our pipeline components. We first evaluate the effectiveness of our geometry-aware reconstruction stage.  The reconstructed coarse 3DGS of LGM often appears blurry and exhibits incorrect textures. To address this, we refine the 3DGS using both depth and color supervision, leading to improved reconstruction quality. As shown in \cref{fig:ablation1}, while the addition of color supervision enhances texture details, it introduces floating artifacts around the object due to geometric ambiguities. Our proposed hard-depth loss mitigates this issue by incorporating depth information, ensuring that the refined front view aligns accurately with the input image and better preserves the object's shape. We then examine the image blending stage. As discussed in \cref{subsec:inversion stage}, applying alpha-blending in the latent space compromises temporal consistency, especially at object edges. \cref{fig:ablation2} shows that inaccuracies in the latent space mask lead to the loss of high-frequency details, such as around the legs of the chair. Lastly, we evaluate the full enhancement pipeline. Prior works \cite{xie2023physgaussian, zhang2025physdreamer} use fully reconstructed 3D scenes to generate physics-based dynamics, serving as coarse dynamics in our pipeline. However, due to challenges in filling internal 3DGS without internal information, \cite{xie2023physgaussian} fails to show realistic dynamics for exposed internal particles, leading to artifacts and blurriness. As shown in \cref{fig:ablation3}, our method enhances blurred sections, e.g. in the bread-tearing video, rendering clear and realistic internal textures.

\section{Conclusion and Future Work}
\label{sec:con}
We introduce PhysMotion, a novel framework that combines physics-integrated 3DGS with image diffusion model to generate physics-grounded dynamics from a single image. Compared with text-guided or trajectory-guided image-to-video methods, PhysMotion achieves more visually realistic and compelling motion synthesis, by enhancing coarse videos obtained from physics simulators through image diffusion models. Nevertheless, the introduction of image diffusion models may introduce additional artifacts, such as slight color distortions, affecting the overall color consistency. How color fidelity can be maintained in video enhancement is still an unsolved problem. Future research can also investigate the integration of video diffusion models with physics-based simulators.

\bibliographystyle{ACM-Reference-Format}
\bibliography{ref}

\newpage
\appendix
\section*{\Large Appendix}
\section{Additional Implementation Details}
In this section we provide comprehensive implementation details.

\subsection{Model Use}
For the pre-trained text-to-image model, we apply the publicly available UNet-based checkpoints of Stable-Diffusion-2-1\footnote{\url{https://huggingface.co/stabilityai/stable-diffusion-2-1}}. We train LoRA weights based on above models for personalization.  For ControlNet models, we utilize public checkpoints for canny-edge-ControlNet\footnote{\url{https://huggingface.co/thibaud/controlnet-sd21-canny-diffusers}} and depth-ControlNet\footnote{\url{https://huggingface.co/thibaud/controlnet-sd21-depth-diffusers}}. We extract the canny-edge control signal using the OpenCV library \cite{opencv_library}, and we adopt the depth map $\mathbf{D}$ from Eq. (2) in the paper or use a depth map extracted from coarse dynamics using MiDaS \cite{stan2023ldm3d}.
\subsection{Parameter Settings}
\subsubsection{Geometry-Aware Reconstruction}
To obtain a 3DGS representation ready to generate reasonable dynamics, our training parameters are carefully chosen: for most of our experiments, the number of training epoch is $3000$, with parameters' learning rates set within the range $[10^{-4},10^{-3}]$.

We apply a learning rate decay strategy by down-scaling the learning rates by $10^{-1}$ after $1500$ epochs; we apply hard-depth supervision every $10$ epochs and after epoch $500$. We do not apply the soft-depth supervision as indicated in \cite{li2024dngaussian} since we do not observe significant change of quality in reconstruction output under our settings.
\subsubsection{Generative Video Enhancement}
We set the deterministic DDIM+ inversion total steps as $1000$, and we set the step size as $20$; following \cite{Tumanyan_2023_Pnp} and \cite{tokenflow2023}, we set the classifier-free-guidance (CFG) \cite{ho2022classifierfree} scale to $1$. We apply deterministic DDIM+ sampling with $50$ steps.

\begin{figure*}[htbp]
 \includegraphics[width=1.0\textwidth]{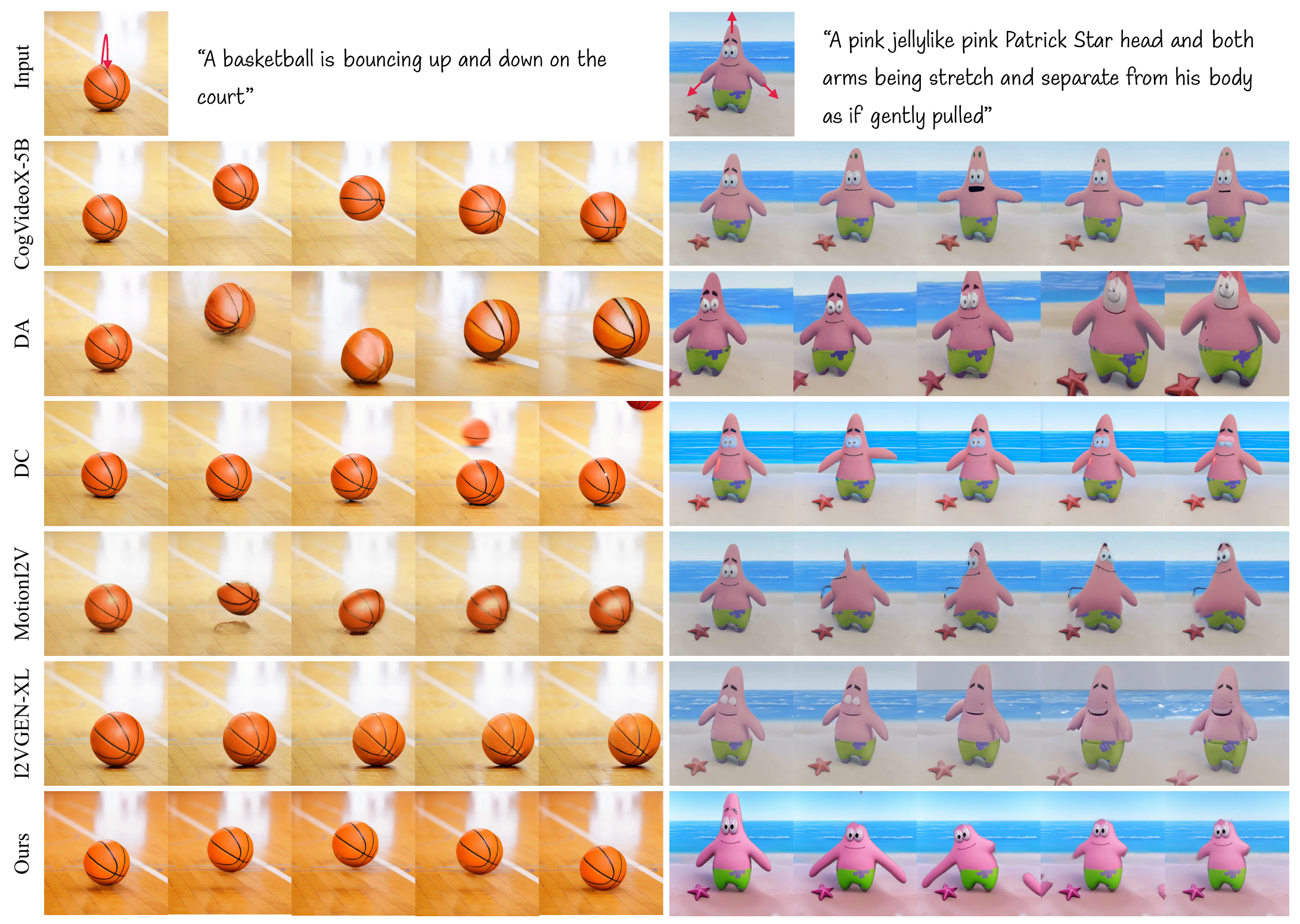}
  \centering
    \caption{\textbf{Additional Qualitative Comparison}.  We provide additional comparison results against MotionI2V \cite{shi2024motion}, DragAnything (DA) \cite{wu2025draganything}, CogVideoX-5B \cite{yang2024cogvideox}, DynamiCrafter (DC) \cite{xing2025dynamicrafter} and I2VGen-XL \cite{shi2024motion}. Text prompts for CogVideoX-5B, I2VGen-XL and DynamiCrafter are generated using ChatGPT-4o, while trajectories are used for DragAnything and Motion-I2V.}
  \label{fig:sup_qua1}
\end{figure*}

We notice that in general, enhanced video achieves better temporal consistency with higher number of key-frames sampled. For most of out experiments, we randomly choose key-frames every $5$ frames, and we set the guidance scale to $7.5$.  
\subsection{Experiment Details}
We observed for different physical scenes in our experiments, VideoPhy \cite{bansal2024videophy} provides scores of different scales in both physics commonsense (PC) and semantic adherence (SA). Therefore, simply calculating the average score is unfair as it does not account for the varying scales of the scores, which could disproportionately influence the results and lead to biased comparisons. To mitigate the influence of heterogeneous scales and ensure a fair comparison across different models, we perform a $z$-score normalization. Specifically, for each scene $t$ ($13$ in total) and score $x_{i,t}$ of model $i$ (chose from \{ours, CogVideoX-5B, DynamiCrafter, I2VGen-XL, MotionI2V, DragAnything\}), we calculate the $z$-score as follows:\begin{equation}
z_{i,t} = \frac{x_{i,t} - \mu_t}{\sigma_t},
\end{equation}
where $\mu_t$ and $\sigma_t$ represent the mean and standard deviation of scores across all models for scene $t$. This normalization allows for a fair comparison across scenes with different scoring scales by transforming scores into a common scale,. We then compute each model’s overall performance by averaging its $z$-scores across all scenes:
\begin{equation}
\Bar{z}_i = \frac{1}{N} \sum_{t=1}^{N} z_{i,t}    
\end{equation}
where $N$ denotes the total number of scenes ($N=13$ in our experiments). Models with higher average $z$-scores demonstrate stronger overall performance across scenes.

\begin{figure*}[htbp]

 \includegraphics[width=\linewidth]{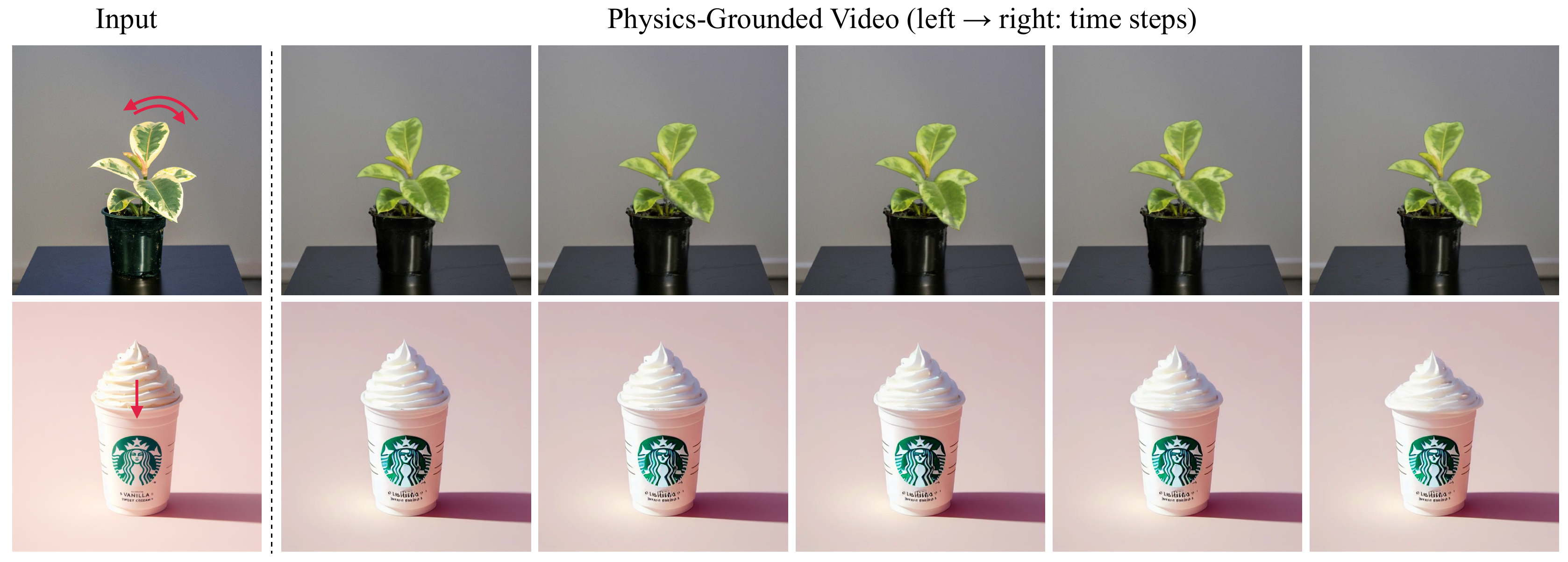}
  \centering
    \caption{\textbf{Additional Showcases.} We demonstrate additional showcases created by PhysMotion.}
  \label{fig:sup_res}
\end{figure*}

\subsection{Additional Preliminary Knowledge}
We provide additional preliminary knowledge on the attention mechanism \cite{Vaswani2017attn}.

In self-attention blocks within transformer blocks, the features $\mathbf{f}\in\mathbb{R}^{n\times d_f}$ ($n$ is the sequence length and $d_f$ is the dimension of feature) are projected into queries $\mathbf{Q}$, keys $\mathbf{K}$, and values $\mathbf{V}$ using\begin{equation}
    \mathbf{Q}=\mathbf{f}\mathcal{W}_{\mathbf{Q}},\quad\mathbf{K}=\mathbf{f}\mathcal{W}_{\mathbf{K}},\quad\mathbf{V}=\mathbf{f}\mathcal{W}_{\mathbf{V}}
\end{equation}
where $\mathcal{W}_{\mathbf{Q}},\mathcal{W}_{\mathbf{K}},\mathcal{W}_{\mathbf{V}}\in\mathbb{R}^{d_f\times d}$ are learned weights matrices for  queries, keys and values respectively. $d$ is the dimension of the embedded vector as in Eq. (12) in the paper.

The attention mechanism computes the weighted sum of the values, with the weights determined by the similarity between queries and keys. Specifically, the attention scores are calculated as the scaled dot product between the queries and keys, as \begin{equation}
    \mathcal{A}=\text{Softmax}\left(\frac{\mathbf{Q}\mathbf{K}^{\mathsf{T}}}{\sqrt{d}}\right),
\end{equation}
where $\mathcal{A}\in\mathbb{R}^{n\times n}$ contains the attention scores for all query-key pairs, with the weighted sum to $1$ for each query. The final output of the attention mechanism is given by\begin{equation}
    \bm\phi = \mathcal{A}\cdot\mathbf{V}.
\end{equation} 
Note that in Eq. (13) in paper, the $\mathbf{V}$'s are concatenated to form a shared value matrix.

\section{More Results}
\subsection{Qualitative Comparison}
In \cref{fig:sup_qua1}, we provide additional qualitative comparison results with baseline methods, including CogVideoX-5B \cite{yang2024cogvideox}, Drag Anything \cite{wu2025draganything}, DynamiCrafter \cite{xing2025dynamicrafter}, Motion-I2V \cite{shi2024motion} and I2VGen-XL \cite{zhang2023i2vgen}.

\subsection{Showcases}
As indicated in \cref{fig:sup_res}, we present additional showcases created using the proposed method. Our method enables users to generate high-fidelity, physics-grounded dynamics.

\clearpage
\end{document}